\documentclass[10pt,twocolumn,letterpaper]{article}

\usepackage{cvpr}              

\usepackage[dvipsnames]{xcolor}

\definecolor{cvprblue}{rgb}{0.21,0.49,0.74}
\usepackage[pagebackref,breaklinks,colorlinks,citecolor=cvprblue]{hyperref}

\def\confName{CVPR}
\def\confYear{2024}

\title{MMA-DFER: MultiModal Adaptation of unimodal models \\ for Dynamic Facial Expression Recognition in-the-wild}

\author{Kateryna Chumachenko\\
Tampere University\\
Tampere, Finland\\
{\tt\small kateryna.chumachenko@tuni.fi}
\and 
Alexandros Iosifidis\\
Aarhus University\\
Aarhus, Denmark\\
{\tt\small ai@ece.au.dk}
\and 
Moncef Gabbouj\\
Tampere University\\
Tampere, Finland
\\
{\tt\small moncef.gabbouj@tuni.fi}}

\begin{document}
{
\maketitle

\begin{abstract}

Dynamic Facial Expression Recognition (DFER) has received significant interest in the recent years dictated by its pivotal role in enabling empathic and human-compatible technologies. Achieving robustness towards in-the-wild data in DFER is particularly important for real-world applications. One of the directions aimed at improving such models is multimodal emotion recognition based on audio and video data. Multimodal learning in DFER increases the model capabilities by leveraging richer, complementary data representations. 
Within the field of multimodal DFER, recent methods have focused on exploiting advances of self-supervised learning (SSL) for pre-training of strong multimodal encoders \cite{hicmae}. Another line of research has focused on adapting pre-trained static models for DFER \cite{s2d}. In this work, we propose a different perspective on the problem and investigate the advancement of multimodal DFER performance by adapting SSL-pre-trained disjoint unimodal encoders. We identify main challenges associated with this task, namely, intra-modality adaptation, cross-modal alignment, and temporal adaptation, and propose solutions to each of them. As a result, we demonstrate improvement over current state-of-the-art on two popular DFER benchmarks, namely DFEW \cite{dfew} and MFAW \cite{mfaw}.

\end{abstract}    
\section{Introduction}
\label{sec:intro}

The ability to perceive non-verbal communication cues is essential for development of truly intelligent interactive technologies. Automatic understanding of human emotional states, whether in the form of facial expressions, voice characteristics, or language semantics, sees a rapid development in the recent years with adoption in a vast number of applications. These include, among others, collaborative robotics \cite{toichoa2021emotion, liu2017facial},  healthcare \cite{bisogni2022impact, ayata2020emotion} and human-computer interaction \cite{chowdary2023deep, zepf2020driver}.

\begin{figure}[h]
\includegraphics[width=8cm]{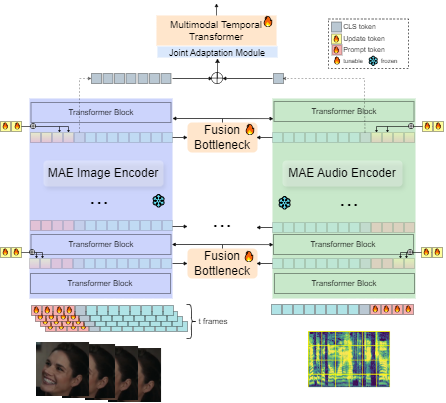}
\caption{Schematic description of MMA-DFER: Two pre-trained frozen MAE encoders are joined by Fusion Bottleneck for modality alignment, followed by joint adaptation module and Multimodal Temporal Transformer. Learnable prompts in each modality independently handle intra-modality gaps between pre-training and downstream data.}
\end{figure}

The prevailing signal for this task has conventionally been the vision modality in the form of facial expressions. The field of facial expression recognition has undergone an evolution from solving static expression recognition problems (SFER) on images captured in controlled environments, to dynamic facial expression recognition (DFER), and further the multimodal dynamic facial expression recognition (most commonly taking the form of audio-visual emotion recognition - AVER). Recently, the field has seen an increased interest towards in-the-wild analysis, facilitated by the introduction of large-scale datasets firstly in static domain \cite{affectnet}, and further in dynamic and multimodal domain \cite{mfaw, dfew}. Methods capable of in-the-wild analysis enable development of novel applications, and ability to handle dynamic scenes increases robustness towards real-world scenarios.

The amount of data surrounding us is vast, and data relevant to facial expression recognition extends beyond the visual domain. Additional data inputs, such as audio or language can aid in the analysis by providing additional information about the signal. This idea has gained momentum also in the DFER field and inspired numerous multimodal facial expression recognition methods. 

Within the multimodal DFER, a large portion of existing works has focused on designing elaborate fusion methods and relying on joint learning of multimodal features that utilizes each modality to its fullest. At the same time, closing the gap between DFER in constrained environments and in-the-wild DFER requires the ability to generalize to wider set of data distributions and to be robust to various challenges and variable factors associated with in-the-wild data. In the context of multimodal DFER, collection of large amounts of multimodal data representative of emotion labels poses a challenging task. 

These facts naturally invite the application of self-supervised learning (SSL) methods, that are able to learn meaningful features from data without requiring labels (followed by downstream fine-tuning for the task in hand). Indeed, previous works aiming at in-the-wild DFER have already shown benefits of such approach in both unimodal \cite{maedfer, sun2023svfap} and multimodal \cite{hicmae} settings. Specifically in the multimodal space, HiCMAE \cite{hicmae} learns a joint audiovisual model by large-scale multimodal SSL pre-training, followed by full fine-tuning for DFER. 

At the same time, an abundance of off-the-shelf unimodal foundational models are publicly available, and progress made in unimodal domains generally precedes their adoption in multimodal space by a significant margin, with multimodal extensions often being ad-hoc and not easily transferable. We find ourselves wishing for universal feature extractors that can be mix-and-matched for multimodal inference without the loss of performance on downstream tasks and without requiring additional multimodal pre-training.  In this work, we aim to address the gap in multimodal adaptation of in-the-wild DFER from only publicly available unimodal foundational models, pre-trained independently, and on unrelated datasets. We show that with appropriate adaptation, we can obtain beyond state-of-the-art results on two popular DFER benchmarks.

Our contributions can be summarized as follows:
\begin{itemize}
\item We show that orthogonally to the works proposed in recent literature, state-of-the-art performance on multimodal dynamic face recognition can be achieved without a) large-scale paired multimodal pre-training; b) pre-training for static facial expression recognition
\item We identify key challenges in adaptation of pre-trained models for multimodal DFER and propose solution to each of them, showing their efficacy. Specifically, this includes progressive prompt tuning for bridging intra-modality gap, Fusion Bottleneck blocks for cross-modal alignment, and Multimodal Temporal Transformer for temporal alignment
    \item We set new SOTA on two popular in-the-wild DFER benchmarks, namely DFEW and MAFW.
\end{itemize}

\section{Related Work}
\label{sec:formatting}
\subsection{Dynamic Facial Expression Recognition in-the-wild}
For a period of time, the field of emotion recognition or facial expression recognition has largely relied on datasets collected in controlled environments and methods designed for them, often leading to suboptimal performance in real-world applications. In the recent years, dynamic facial expression recognition in-the-wild has emerged as a separate task within the paradigm of affective behavior analysis, thanks to the appearance of a number of large-scale in-the wild datasets \cite{dfew, mfaw, abaw1}. Recently, a number of novel methods has emerged in this area. Among them, there is an ongoing trend of reliance on large-scale models performing large-scale pre-training, often in a self-supervised manner. Specifically, DFER-CLIP \cite{dferclip} relies on joint text-image space of CLIP for mapping dynamic videos to emotion labels; SVFAP \cite{sun2023svfap} and MAE-DFER \cite{maedfer} explore Video-MAE-like masked reconstruction pre-training on dataset of face videos; HiCMAE \cite{hicmae} extends this idea to multimodal inputs. Our work differs from these approaches in a way that we explore adaptation of pre-trained unimodal models for multimodal DFER without multimodal pre-training.

Another line of work, most similar to ours in spirit, is S2D \cite{s2d} that explores temporal expansion of pre-trained static facial expression recognition model for dynamic recognition, and employs temporal adaptation modules, and landmark-guided modules to achieve competitive results. We pose a similar question in the multimodal domain, and show that we can outperform S2D with 20\% less trainable parameters (both ours and S2D vision encoder follows ViT-b architecture), while accommodating another modality and not requiring static facial expression recognition pre-training, only self-supervised pre-training in each modality independently.

\subsection{Adaptation of pre-trained models}

End-to-end fine-tuning, or fine-tuning with a frozen backbone has been a dominant approach in the task of adaptation of large-scale pre-trained models for downstream tasks. Recently, dictated by the adoption of Large Language Models, a different paradigm emerged, namely prompt tuning, primarily targeted at Transformer models. In the NLP domain, prompt tuning was introduced as a way to adapt the model for a downstream task, while requiring substantially less trainable parameters by introduction of additional tokens in the input space, that are concatenated to the input, but are optimized via backpropagation \cite{li2021prefix, liu2023pre, liu2021p}. Visual prompt tuning, introduced in \cite{jia2022visual} extends this idea to the vision domain, where learnable tokens are concatenated to the input sequence of patches of a pre-trained model. A few works aim at cross-modal adaptation via prompt tuning \cite{khattak2023maple} (targeted at VLMs), where prompts are introduced in multiple input modalities, and interact throughout the model.

While tackling a multimodal task, our adoption of prompt tuning is aimed at each modality independently, with the goal of reducing the domain shifts in each of the unimodal foundation models. Moreover, we introduce the idea of progressive prompt adaptation, aimed at adapting the model at multiple levels of granularity independently, and therefore simplifying the adaptation task.

Another challenge differentiating DFER from static facial expression recognition from images is the need to learn temporal dependencies in the data, and identify most relevant ones. To address this challenge, prior works have either employed temporally-aware pre-training on videos \cite{hicmae, maedfer} or performed temporal adaptation of pre-trained static models at the fine-tuning stage \cite{s2d, dferclip}. A natural choice for the latter approach is often a form of temporal self-attention, introduced in different stages of feature extraction. Working with a static image encoder, we also follow a similar temporal self-attention-based approach and propose a Multimodal Temporal Transformer for temporal information extraction. We also  evaluate its efficacy at different stages of the model.

In the field of multimodal facial expression recognition, prior works have to a large extent focused on joint cross-modal feature learning through complex fusion methods and/or multimodal pre-training \cite{hicmae, chumachenko2022self, ahmed2023systematic, pan2023review, tellamekala2023cold, mocanu2023multimodal}. The fusion methods are often ad-hoc and range from simple concatenation, to multimodal fusion Transformers or other elaborate frameworks. We are instead focusing on adapting already pre-trained unimodal models via lightweight bottleneck fusion adaptors while preserving their unimodal feature extraction capabilities. 
\section{Methodology}

Our method operates by employing two off-the-shelf SSL-pre-trained unimodal encoders trained independently on audio sequences and static images, and adapts them for audiovisual dynamic facial expression recognition in-the-wild. Specifically, we choose among the models pre-trained with Masked Autoendoder (MAE) reconstruction objective, dictated by outstanding performance of MAEs on tasks requiring fine-grained features \cite{mae}. Following this, as foundation models we employ two publicly available Vision Transformer-based encoders, namely AudioMAE \cite{audiomae} and MAE-Face \cite{maeface} that both follow ViT-base \cite{vit} architecture and therefore have the same depth. AudioMAE is pre-trained on AudioSet \cite{audioset} and MAE-Face is pre-trained on a combination of static face image datasets \cite{maeface}.

While large-scale pre-training of unimodal encoders ensures rich and to an extent generalizable representation within each modality, it is associated with certain challenges when adapting to multimodal DFER. These challenges can be distinguished into three categories:
\begin{itemize}
    \item \textbf{Within-modality domain gap}: naturally, although pre-trained on large-scale data, certain level of domain gap can be expected between the data distributions of the pre-trained models and the downstream domains, especially if the pre-training datasets have little relevance to DFER.
    \item \textbf{Modality alignment gap}: considering a model where two encoders are pre-trained without awareness of another modality, and potentially on different data sets, there exists a gap between the distributions of feature spaces of the two modalities, both on the decision-level, and on the intermediate feature level.
    \item{\textbf{Temporal adaptation gap}}: considering a pre-trained model that was trained on static images, adaptation to dynamic facial expression recognition could benefit from learning temporal dependencies between the frames.
\end{itemize}

\noindent
In the following sections, we discuss how to address each of these gaps.

\subsection{Formulation}
Let us start with formally defining a problem. Each data sample is represented by a pair of audio sequence and frame sequence and corresponds to a single emotion class label. Each frame is initially split into patches independently, following the Transformer embedding layer \cite{vit, vaswani2017attention}. Frames corresponding to the same video are concatenated batch-wise and we describe their temporal interaction within the batch further. Similarly, a spectrogram is extracted from audio sequence and split into patches, too, following the AudioMAE \cite{audiomae} procedure.

\subsection{Progressive prompt learning}

First, we propose to address the domain gap within each modality encoder independently. To achieve this, we introduce a set of learnable prompts for each modality that are concatenated to the data sequences and are updated via backpropagation. As the tokens are processed by the model, learnable prompts interact with the data tokens and are able to divert their feature representation distribution so that it is closer to the initial distribution of the data on which the model was trained. 

Specifically, learnable prompts are a set of $M$ tokens $\mathbf{p}^m$ in $\mathcal{R}^d$, where $d$ is a hidden dimension of the encoder, and the learnable prompts are concatenated to the input sequence following the embedding layer. 

We note that Transformer adaptation via learnable prompts has been shown to be successful in few application areas. However, in the prior work, this idea is applied purely on the input space level. At the same time, discrepancies in feature distributions at different depths can have different nature, and it can be difficult to address all of them with the same prompt on the input level. Instead, applying dedicated prompts at different depths can aid the adaptation. Therefore, we propose \textit{progressive prompt adaptation}, where we introduce a set of $M^l$ tokens at different depths of the model, that complement the initial $M$ learnable prompt tokens and are introduced to the network progressively. Formally, given $M$ learnable prompt tokens $\mathbf{P} \in \mathcal{R}^{M \times d}$, and $L$ intermediate layers, we introduce $L$ sets of tokens $\mathbf{P}^l \in \mathcal{R}^{M^l \times d}$, $M^l = \frac{M}{L}$ and at $l^{th}$ layer, the prompts are updated as:
\begin{equation}
    \mathbf{P}[(l-1)*M^l:l*M^l,:] = \mathbf{P}[(l-1)*M^l:l*M^l:] + \mathbf{P}^l.
\end{equation}

This way, adaptation of the model to new data distribution happens progressively thereby simplifying the task.

\subsection{Modality fusion bottlenecks}

To exploit cross-modal dependencies for knowledge extraction, we introduce fusion of modalities at multiple intermediate stages. Our Modality Fusion blocks are based on creating a low-dimensional bottleneck where multimodal features are obtained by fusing compressed and normalized representations of each modality, and routing them back to the corresponding unimodal branches via a gating mechanism. This approach relates to principles of learning to control the amount of transferred information, as well as information compression followed by its expansion for highlighting the most relevant features, that have been successfully applied as building blocks of different methods in the field  \cite{rezero, flamingo, ae, squeeze}. Our formalization in modality fusion allows to flexibly adapt pre-trained unimodal models. A schematic representation is provided in the Figure \ref{bottleneck}.

Given a video representation $\mathbf{V} = \{\mathbf{v}\}_t$ corresponding to $t$ frames, and an audio sequence $\mathbf{A}$, we first project each of them to a low-dimensional latent space to obtain corresponding low-dimensional representations $\hat{\mathbf{V}}$ and $\hat{\mathbf{A}}$:
\begin{equation}
\centering
\begin{split}
    \hat{\mathbf{V}} & = \{\mathcal{G}_V(\mathbf{v})\}_t,\\
    \hat{\mathbf{A}} & = \mathcal{G}_A(\mathbf{A}),
    \end{split}
\end{equation}
where $\mathcal{G}_V$ and $\mathcal{G}_A$ consist of a Linear layer projecting from high- to low-dimensional space, and a Layer Normalization layer. Further, each low-dimensional representation is followed by a corresponding aggregation function to obtain global sequence representation in each modality. For audio modality, this is represented by a Global Average Pooling over tokens of audio sequence, and for vision modality, by Global Average Pooling over tokens of all image sequences in all frames corresponding to one video:
\begin{equation}
    \mathbf{L}_v = \frac{\sum_{i=1}^t\sum_{j=1}^{N_v}\hat{\mathbf{V}}_{i}^{j}}{t*N_v} \hspace{0.5cm} \textrm{and} \hspace{0.5cm}
    \mathbf{L}_a = \frac{\sum_{j=1}^{N_a}\hat{\mathbf{A}}^{j}}{N_a}.
\end{equation}

\begin{figure}[h]
\includegraphics[width=8cm]{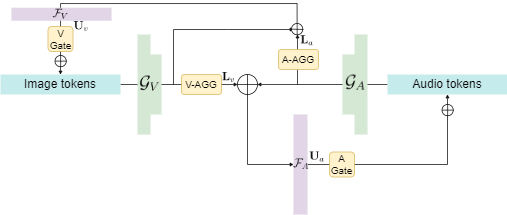}
\caption{Fusion bottleneck. The features of each modality are first compressed to a low-dimensional representation, aggregated to obtain global representation per sequence, then fused with the complementary modality, and the joint representation is expanded back to the original space, and added to the initial features via a gating mechanism.}
\label{bottleneck}
\end{figure}

Having obtained global low-dimensional representations for each modality, we fuse them via addition to the opposite modality (unaggregated), and employ upsampling functions $\mathcal{F}_A$ and $\mathcal{F}_V$ to expand the joint representations back to the original dimensionality spaces. In our implementation, $\mathcal{F}_A$ and $\mathcal{F}_V$ are represented by a Linear layer, followed by a GELU activation:
\begin{equation}
\begin{split}
    \mathbf{U}_a & =\mathcal{F}_A(\hat{\mathbf{A}} + \mathbf{L}_{v}), \\
    \mathbf{U}_v & =\mathcal{F}_V(\hat{\mathbf{V}} + \mathbf{L}_{a}). 
\end{split}
\end{equation}

Finally, the obtained fused representations are added to the original ones via gated skip-connections, where we employ a learnable parameter $\alpha$ that controls the strength of multimodal representations:
\begin{equation}
\begin{split}
    \tilde{\mathbf{A}} & = \mathbf{A} + \tanh(\alpha)*\mathbf{U}_a, \\
    \tilde{\mathbf{V}} & = \mathbf{V} + \tanh(\alpha)*\mathbf{U}_v.
\end{split}
\end{equation}
We initialize $\alpha$ to zero such that the modality encoders receive the original (unimodal) feature representations in the first iterations of the training, and the appropriate magnitude of multimodal features is progressively learnt by the model, hence making adaptation to multimodal features easier.

As a result, compression of the features of each modality to a low-dimensional latent space helps highlight the most relevant information in each modality, and further expand it back to the corresponding branch, while keeping this process adaptive to the strength of each modality via gating.

\subsection{Temporal fusion}
The task of dynamic facial expression recognition can benefit from knowledge about between-frame temporal dependencies. However, a model that was pre-trained on static images lacks such capability. We address this challenge by employing a Multimodal Temporal Transformer. 

Recall that in the image encoder, given a sequence of $t$ frames of a video, we concatenate them in a batch manner and process independently in a parallel manner. Following this stage, we extract the $[CLS]$ token of every frame corresponding to the same video and concatenate them to form a temporal sequence. Therefore, the input tensor in the vision branch is reshaped from $B*t \times N \times D$ to $B \times t \times D$, where $B$ is the batch size, $t$ is the temporal length (number of frames), $N$ is the sequence length of each frame (number of patches), and $D$ is the dimensionality. We additionally fuse the corresponding $[CLS]$ token of audio branch to the video sequence via addition, and process the new multimodal sequence with a Joint Adaptation Module, which in our implementation is formalized as a single Linear layer (with slightly reduced dimensionality than prior encoders). Further, we add learnable temporal embeddings to the multimodal temporal sequence and concatenate a new $[CLS]$ token. We process the new sequence with a Transformer block, which is now operating on temporal level on joint multimodal sequence, hence we refer to it as Multimodal Temporal Transformer. The $[CLS]$ token is further used as input to the classifier.

We also note that some works have suggested adoption of temporal adaptation modules on intermediate feature level \cite{s2d}, with the aim of enriching image features with temporal information already at the feature extraction stage. We perform ablation studies on the placement of our Multimodal Temporal Transformer with intermediate temporal modules in experimental section, and find our approach to be beneficial. This can be partially attributed to fusion modules that already implicitly embed a certain level of temporal information at intermediate stages by fusing an audio sequence to image patches.

\section{Experiments}

\subsection{Datasets}
We benchmark our method on two popular multimodal dynamic facial expression recognition in-the-wild datasets, namely DFEW \cite{dfew} and MAFW \cite{mfaw}. DFEW consists of 16,000 audiovisual clips split into 5 folds, posing a 7-class classification task, with classes being emotion labels of `happy', `sad', `angry', `neutral', `surprise', `disgust', and `fear'. MAFW contains 10,045 clips and follows a 5-fold split as well. In this dataset set of emotions follows 11 classes: `anger', `disgust', `fear', `happiness', `sadness', `surprise', `contempt', `anxiety', `helplessness', `disappointment', and `neutral'. 

\subsection{Experimental setup}
 
We follow the traditional setup of extracting 16 frames to form a video sequence during training \cite{s2d, hicmae}. We notice the discrepancies in image sizes in the experimental setup of existing methods, therefore report results on 112x112 images \cite{mfaw}, 160x160 images \cite{hicmae}, and on 224x224 images \cite{s2d}. For MAFW, we extract faces with MTCNN \cite{mtcnn}. We use learning rate of 1e-4 that is annealed via cosine schedule to 0 over 25 epochs. We use batch size of 8 and weight decay of 1e-2, and AdamW optimizer with default parameters. We fix the random seed to 1. In unimodal encoders, we interpolate the positional embeddings to new sequence lengths and keep them tunable to adapt to the new resolutions, the rest of unimodal encoder parameters remain frozen. Therefore, tunable parameters include the Fusion Bottleneck blocks, Multimodal Temporal Transformer, learnable prompts, classifier and positional embeddings, totaling ~7.5M parameters. For comparison, S2D that follows a similar framework of adapting pre-trained (for static emotion recognition in their case) network, includes 9M tunable parameters.

Multimodal Temporal Transformer is a 1-layer Transformer with hidden dimension of 512 and 8 heads. Fusion Bottleneck latent dimensionality is set to 128, and 6 learnable prompts are introduced in each modality, and progressive updates are introduced twice in the network, after 1st and 7th layers, with 3 tokens each.

We follow the 5-fold experimental protocol in each dataset with the provided splits. We train the models on the train set and report the result of final checkpoint, i.e., at 25th epoch, on the test set. Training is done on a single Nvidia Tesla V-100-32 GPU, and single training (of single fold) takes approximately 8 hours on resolution 160. We report results both with following the 16-frame uniform sampling at inference time, as well as widely adopted 2-clip average results \cite{s2d, hicmae}, where two clips are sampled uniformly from a single video and predictions are averaged. We note that we do not observe a significant difference between these two approaches. As prior works, we report Unweighted Average Recall (UAR) and Weighted Average Recall (WAR).

\subsection{Results}

\begin{table}[]
\centering
\small
\setlength{\tabcolsep}{4pt}
\begin{tabular}{lcccccc}
\hline
                & \multicolumn{2}{c}{DFEW}                          & \multicolumn{2}{c}{MAFW}                          \\ \hline
Method          & UAR & WAR & UAR & WAR & M & Res. \\
\hline
Wav2Vec2.0 \cite{wav2vec} & 36.15&43.05&21.59&29.69&A&-\\
HuBERT \cite{hubert}&35.98&43.24&25.00&32.60&A&-\\
WavLM-Plus \cite{wavlmplus} &37.78&44.64&26.33&34.07&A&-\\
C3D+LSTM \cite{mfaw} &53.77&65.17&30.47& 44.15 & AV & 224 \\
T-ESFL \cite{mfaw} &-&-&33.35 & 48.70 & AV & 224 \\

C3D \cite{c3d}           & 42.74                   & 53.54                   & 31.17                   & 42.25   & V & 112                \\
R(2+1)D-18  \cite{6}    & 42.79                   & 53.22                   & -                       & -           & V & 112           \\
3D ResNet-18 \cite{7}   & 46.52                   & 58.27                   & -                       & -         & V & 112              \\
Former-DFER  \cite{8}   & 53.69                   & 65.70                   & -                       & -     & V & 112                  \\
CEFLNet  \cite{9}       & 51.14                   & 65.35                   & -                       & -      & V & 224                 \\
T-ESFL   \cite{mfaw}       & -                       & -                       & 33.28                   & 48.18        & V & 224           \\
EST   \cite{10}          & 53.43                   & 65.85                   & -                       & -        & V & 224               \\
STT   \cite{23}          & 54.58                   & 66.65                   & -                       & -            & V & 112           \\
NR-DFERNet   \cite{66}   & 54.21                   & 68.19                   & -                       & -        & V & 112               \\
AMH \cite{amh} & 54.48 & 66.51 & 32.98 & 48.83 & AV & 224\\
IAL   \cite{11}          & 55.71                   & 69.24                   & -                       & -               & V & 112        \\
M3DFEL   \cite{51}       & 56.10                   & 69.25                   & -                       & -            & V & -           \\
CLIPER   \cite{12}       & 57.56                   & 70.84                   & -                       & -  & V & 224                     \\
TMEP \cite{tmep} & 57.16 & 68.85 &37.17 & 51.15 & AV & 112\\
DFER-CLIP  \cite{13}     & 59.61                   & 71.25                   & 38.89                   & 52.55 & V &   224               \\
SVFAP \cite{sun2023svfap} & 62.83 & 74.27 & 41.19 & 54.28 & V & 160 \\
MAE-DFER   \cite{maedfer}     & 63.41                   & 74.43                   & 41.62                   & 54.31     & V & 160              \\
HiCMAE   \cite{hicmae}       &              63.76            &            75.01             & 42.65                        &       56.17   & AV & 160               \\
S2D   \cite{s2d}          & 65.45                   & 76.03                   & 43.40                   & 57.37 & V & 224                  \\ \hline
MMA (ours)\_112 & 64.24 & 75.40 & 43.34 & 56.46 & AV & 112 \\
MMA (ours)\_112* & 64.35 & 75.48 & 43.29& 56.60 & AV & 112 \\
MMA (ours)\_160 & 66.61    & 77.15    &  44.19   & 57.90  & AV & 160  \\
MMA (ours)\_160* & 66.51    & 77.10   & 44.07    &57.85 & AV & 160  \\
MMA (ours)\_224 & \textbf{67.01}    & \textbf{77.51}    & 44.11   &\textbf{58.52} & AV & 224    \\
MMA (ours)\_224* & 66.85   & 77.43    & \textbf{44.25}   & 58.45  & AV & 224  \\
 \hline
\end{tabular}
\caption{Comparison to SOTA methods. * denotes mean prediction over two uniformly sampled video clips following \cite{s2d, hicmae}. M denotes the modality, and Res denotes the image resolution. }
\label{sota}
\end{table}

The comparison to state-of-the-art methods is provided in Table~\ref{sota}. As can be seen, MMA-DFER outperforms the competing methods. Specifically, MMA-DFER outperforms current state-of-the-art of S2D by 1.5\% UAR and WAR on DFEW dataset and 1\% on MAFW. We also note that for S2D, the best UAR and WAR are obtained from different models/training strategies (with and without oversampling of underrepresented classes), while in MMA-DFER this is achieved by a single model. Compared to the best multimodal model - HiCMAE, we achieve stronger results as well, both on 224 and 160 resolution. With the same image resolution, we obtain 2-3\% improvement on DFEW and 1.5\% on MAFW.

\subsection{Comparison of temporal adaptation approaches}

In MMA, the Multimodal Temporal Transformer is placed following the unimodal branches. However, some works have suggested a different approach, where temporal adaptation happens already in the intermediate features of the model \cite{s2d}. We perform an experiment where we add such temporal adaptors in the intermediate steps of the network, following each Bottleneck Fusion block. Such temporal adaptors are similar to TMA described in \cite{s2d} and in practice we apply temporal self-attention over $[CLS]$ tokens of every frame and fuse the obtained representation back to the vision branch. We evaluate intermediate temporal adaptors (ITA), Multimodal Temporal Transformer (MTM), and the combination of the two. In the case where only intermediate temporal adaptors are applied, we control for the total number of parameters and therefore each intermediate block is smaller than our Multimodal Temporal Transformer. We also provide results with different dimensionalities of intermediate temporal adaptors. We compare the results on 1st fold of DFEW on 160x160 resolution.

The results can be seen in Table~\ref{TMA}, where we report UAR, WAR, and number of parameters that correspond to temporal processing blocks. We note that the variant with ITA of dimensionality 128 has less parameters than ITA-64 due to internal dimensionality of Bottleneck Fusion already being 128, therefore no additional layers need to be introduced to this variant to project to new dimensionality space, unlike all other variants. We can see that the best result is achieved by MTM and the second-best by MTM combined with ITA. Among ITA, we observe the variant with d=128 to be outperforming the competing ones. We also note that the performance is not directly correlated with the number of parameters, but more with the placement of the temporal module.

\begin{table}[]
\begin{tabular}{lccc}
\hline
\multicolumn{1}{l}{Temporal adaptation} & \multicolumn{1}{l}{UAR} & \multicolumn{1}{l}{WAR} & Params (M) \\ \hline
ITA, d=64                        &               63.27           &         76.59              & 1.3   \\
ITA, d=128                       &             64.44             &        76.80   & 0.8              \\
ITA, d=256                       &             63.35             &           76.59  & 4.3\\
MTM + ITA-128                                     &        64.66                  &      77.36             & 2.9     \\
MTM (ours) &\textbf{ 66.52} & \textbf{77.92} & 2.2\\
\end{tabular}
\caption{Comparison of Multimodal Temporal Transformer vs intermediate temporal blocks}
\label{TMA}
\end{table}

\subsection{Comparison of fusion approaches}

We additionally compare our Fusion Bottleneck blocks with other popular multimodal fusion approaches. Specifically, we compare with:
\begin{itemize}
    \item Addition in original space, i.e., without compression and expansion (ADD);
    \item Multimodal Transformer following \cite{mult}, where we introduce 2 multi-head self-attention blocks, audio-to-vision, and vision-to-audio, outputs of which are added to corresponding branches (MULT); 
    \item Single multimodal Transformer on concatenated audio and image features, with global average pooled representation added back to each branch (MULT-concat).
    \item No modality fusion;
\end{itemize}
For MULT, we utilize 2 heads with total dimensionality of 128, to match the Fusion Bottleneck dimensionality. We compare the results on 1st fold of DFEW on 160x160 resolution.

The results can be seen in Table~\ref{fusion}. As can be seen, our approach outperforms the competing methods by a significant margin, indicating the effectiveness of the proposed Fusion Bottlenecks. Poor performance of MULT and MULT-concat can be associated with difficulty of drawing dependencies between individual frames and whole audio spectrogram. 

\begin{table}[]
\centering
\begin{tabular}{lcc}
\hline
\multicolumn{1}{l}{fusion methods} & \multicolumn{1}{l}{UAR} & \multicolumn{1}{l}{WAR} \\ \hline
None                           &       59.27                   &            70.56              \\
MULT                                 &     59.89                     &         71.12                 \\
MULT-concat                          &     54.96                     &     65.66                     \\
ADD & 59.97 & 71.20\\
Fusion Bottleneck (ours)                              &            \textbf{66.52}              &           \textbf{77.92  }            
\end{tabular}
\caption{Comparison of modality fusion approaches.}
\label{fusion}
\end{table}

\subsection{Ablation studies}

Further, we ablate each component of our model independently, with the results shown in Table~\ref{ablation}. Here, `Pr.' corresponds to learnable prompts that are not updated with depth but the initialization is kept; `Pr.Pr.' corresponds to the alternative with the progressive prompts, i.e., new learnable prompts are introduced at different depths; `MTT' corresponds to Multimodal Temporal Transformer, and `FB' corresponds to Fusion Bottleneck. When no MTT is employed, prediction is done on averaged features following Joint Adaptation Module. Joint Adaptation Module and a classifier are present and unfrozen in all variants. We report the results on first fold of DFEW.

\begin{table}[t]
\centering
\begin{tabular}{cccccc}
\hline
\multicolumn{1}{c}{Pr} & \multicolumn{1}{c}{Pr.Pr.} & \multicolumn{1}{c}{MTT} & \multicolumn{1}{c}{FB} & \multicolumn{1}{c}{UAR} & \multicolumn{1}{c}{WAR} \\ \hline
-                             & -                                        & -                                    & -                                  &               56.59           &        68.43                  \\
+                             & -                                        & -                                    & -                                  &            57.45             &            69.33              \\
+                             & +                                        & -                                    & -                                  &          58.21             &         69.34                 \\
+                             & +                                        & +                                    & -                                  &            59.27              &             70.56             \\
+                             & +                                        & -                                    & +                                  &  61.96                        &   74.58                       \\
-                             & -                                        & +                                    & +                                  &       64.62                   &        77.40                  \\
+                             & +                                        & +                                    & +                                  &           \textbf{66.52}               &          \textbf{77.92}               
\end{tabular}
\caption{Ablation studies}
\label{ablation}
\end{table}
As can be seen, performance of plain pre-trained models is rather poor, and each of the components progressively improves the performance. The biggest effect is brought by Fusion Bottleneck which improves WAR by 5\% and UAR by 3.7\% if comparing variants without MTT. We also note how performance increases by using MTT by 1.5\%, but when used in conjunction with Fusion Bottlenecks, MTT improves the performance by 3.5\%. This shows that fusion of features on intermediate level helps late-stage adaptation that precedes temporal modeling.

We further study the effect of each modality independently. Here, in the unimodal cases each branch is frozen and only the classifier is updated. The results are shown in Table~\ref{unimodal}. As can be seen, performance of each individual modality is significantly lower than the combination. 

\begin{table}[]
\centering
\begin{tabular}{lcc}
\hline
            & UAR            & WAR            \\
            \hline
Audio only  & 26.43          & 33.53          \\
Vision only & 54.34          & 67.15          \\
MMA-DFER    & \textbf{66.52} & \textbf{77.92}
\end{tabular}
\caption{Comparison of each modality encoder to the multimodal MMA-DFER}
\label{unimodal}
\end{table}

We additionally study the dimensionality of the bottleneck space. Recall that dimensionality of both audio and image encoders in ViT-base are 768 \cite{vit}, and we set our bottleneck space to 128. Here, we evaluate the effect of larger and smaller dimensionalities of the latent space. The results are shown in Table~\ref{dimensions}. As can be seen, larger values do increase the performance of the model, but require significantly more parameters. We also find that increasing dimensionality beyond d-128 brings rather diminishing returns in terms of UAR and WAR.

\begin{table}[]
\centering
\begin{tabular}{lccc}
\hline
\multicolumn{1}{l}{latent space d} & \multicolumn{1}{l}{UAR} & \multicolumn{1}{l}{WAR} & Params (M) \\ \hline
64                                   &   64.57                      &     77.44              &   5.9    \\
128                                  &    66.52                      &      77.92              &     7.5 \\
256                                  &    66.72                    &       \textbf{78.17}                  & 12.3\\
512                                  &    \textbf{66.77}                      &          78.13 & 21.7
\end{tabular}
\caption{Dimensionality of the latent space}
\label{dimensions}
\end{table}

Finally, we also study the frequency of prompt updates in the case of progressive prompting. The results are shown in Table~\ref{progr}. Here, 12 corresponds to introducing new learnable prompts every layer, 6 every 2nd layer, 2 every 3rd layer, 2 every 6th layer (our final model case), and 0 not using learnable prompts at all. As can be seen, any number of updates results in better accuracy than no prompts, with the best result achieved when introducing them twice.

\begin{table}[]
\centering
\begin{tabular}{lll}
\hline
\multicolumn{1}{l}{Pr.Pr. Num.} & \multicolumn{1}{l}{UAR} & \multicolumn{1}{l}{WAR} \\ \hline
12                                         &      65.14                    &             77.87             \\
6                                          &      65.44                  &         77.83               \\
4                                          &         64.41                 &            77.62              \\
2                                          &        \textbf{66.52}                  &             \textbf{77.92}           \\
0 & 64.62 & 77.40 \\
\end{tabular}
\caption{Ablation on the frequency of progressive prompts}
\label{progr}
\end{table}

\section{Conclusion}
We have investigated adaptation of pre-trained unimodal models for multimodal dynamic facial expression recognition in-the-wild. We identified key limitations associated with adapting pre-trained models for this task, namely intra-modality adaptation, cross-modal alignment, and temporal adaptation, and proposed solutions to address them. Our proposed model, MMA-DFER sets a new state-of-the art on two popular DFER benchmarks DFEW and MAFW. Future work could include experimentation with additional unimodal backbones and exploitation of further modalities/sensors, such as landmarks or vision-language latent spaces.

\section{Acknowledgement}
This work was partially funded by the NSF CBL and Business Finland project AMALIA, and by the Horizon Europe programme PANDORA (GA 101135775).
}
{
    \small
    \bibliographystyle{ieeenat_fullname}
    \bibliography{main}
}

\end{document}